\documentclass{article}
\usepackage{amsmath,graphicx,02460}
\usepackage{hyperref}
\usepackage{enumitem}
\usepackage{multirow}

\toappear{Accepted in \textit{Journal of Pathology Informatics}}
\title{Automated detection of celiac disease on duodenal biopsy slides:\\ a deep learning approach}
\name{\normalsize \normalfont Jason W. Wei,$^{1,2}$\sthanks{jason.20@dartmouth.edu} Jerry W. Wei,$^{1}$ Christopher R. Jackson,$^{3}$ Bing Ren,$^{3}$ Arief A. Suriawinata,$^{3}$ Saeed Hassanpour$^{1,2,4}$\sthanks{saeed.hassanpour@dartmouth.edu}}
\address{\normalsize  $^{1}$Department of Biomedical Data Science, Dartmouth College \\
\normalsize  $^{2}$Department of Computer Science, Dartmouth College\\ 
\normalsize  $^{3}$Department of Pathology and Laboratory Medicine, Dartmouth-Hitchcock Medical Center\\
\normalsize  $^{4}$Department of Epidemiology, Dartmouth College\\}

\begin{document}
%

\maketitle
\begin{abstract}
Celiac disease prevalence and diagnosis have increased substantially in recent years. The current gold standard for celiac disease confirmation is visual examination of duodenal mucosal biopsies. An accurate computer-aided biopsy analysis system using deep learning can help pathologists diagnose celiac disease more efficiently. 
In this study, we trained a deep learning model to detect celiac disease on duodenal biopsy images. Our model uses a state-of-the-art residual convolutional neural network to evaluate patches of duodenal tissue and then aggregates those predictions for whole-slide classification. We tested the model on an independent set of 212 images and evaluated its classification results against reference standards established by pathologists. 
Our model identified celiac disease, normal tissue, and nonspecific duodenitis with accuracies of 95.3\%, 91.0\%, and 89.2\%, respectively. The area under the receiver operating characteristic curve was greater than 0.95 for all classes. 
We have developed an automated biopsy analysis system that achieves high performance in detecting celiac disease on biopsy slides. Our system can highlight areas of interest and provide preliminary classification of duodenal biopsies prior to review by pathologists. This technology has great potential for improving the accuracy and efficiency of celiac disease diagnosis. 

\end{abstract}
\begin{keywords}
celiac disease, biopsy, whole-slide imaging, deep learning, digital pathology
\end{keywords}
\section{Introduction}
Celiac disease (CD), an autoimmune disorder triggered from the consumption of gluten, affects as much as one percent of the population worldwide [1, 2]. Patients who are diagnosed with CD undergo treatment in the form of a lifelong gluten-free diet, which requires substantial patient education, motivation, and follow-up [3]. Recent studies have found that the prevalence of CD has increased dramatically in the United States and Europe, and that undiagnosed CD was associated with a nearly four-fold increase in risk of death [4-6]. In fact, CD remains undiagnosed in the majority of affected people, highlighting the need for more frequent and accurate methods for its detection [7-9].

Celiac disease diagnosis involves serological testing of celiac-specific antibodies, followed by microscopic examination of duodenal biopsies, which are considered the gold standard in diagnostic confirmation of CD [10, 11]. Typically, four to six duodenal samples are taken from the patient by an endoscopic procedure, and these samples are visually examined by a pathologist. A confirmatory diagnosis requires detection of histological changes associated with the disease, which are classified according to guidelines from either Marsh [12], Marsh modified (Oberhuber) [13], or Corazza [14]. Endoscopic findings that indicate CD include scalloped folds with or without mosaic pattern mucosa, reduction in the number of folds, and nodular mucosa [15]. The spectrum of histologic changes in CD ranges from only increasing intraepithelial lymphocytes with preserved villous architecture to mild villous blunting to complete villous atrophy [16]. Studies have shown that the histological diagnosis of biopsies are subject to a significant degree of inter-observer variability [17-20]. One potential method for improving the accuracy of CD detection on duodenal biopsies is to apply automated image analysis to aid pathologists. Because the prevalence of CD is increasing, active case-finding is currently being used to screen more patients [21, 22]. An automated biopsy analysis system could help pathologists by filtering and pre-populating scans, improving efficiency and turnaround-time. 

\begin{figure*}[ht]
\begin{centering}
\includegraphics[width=\linewidth]{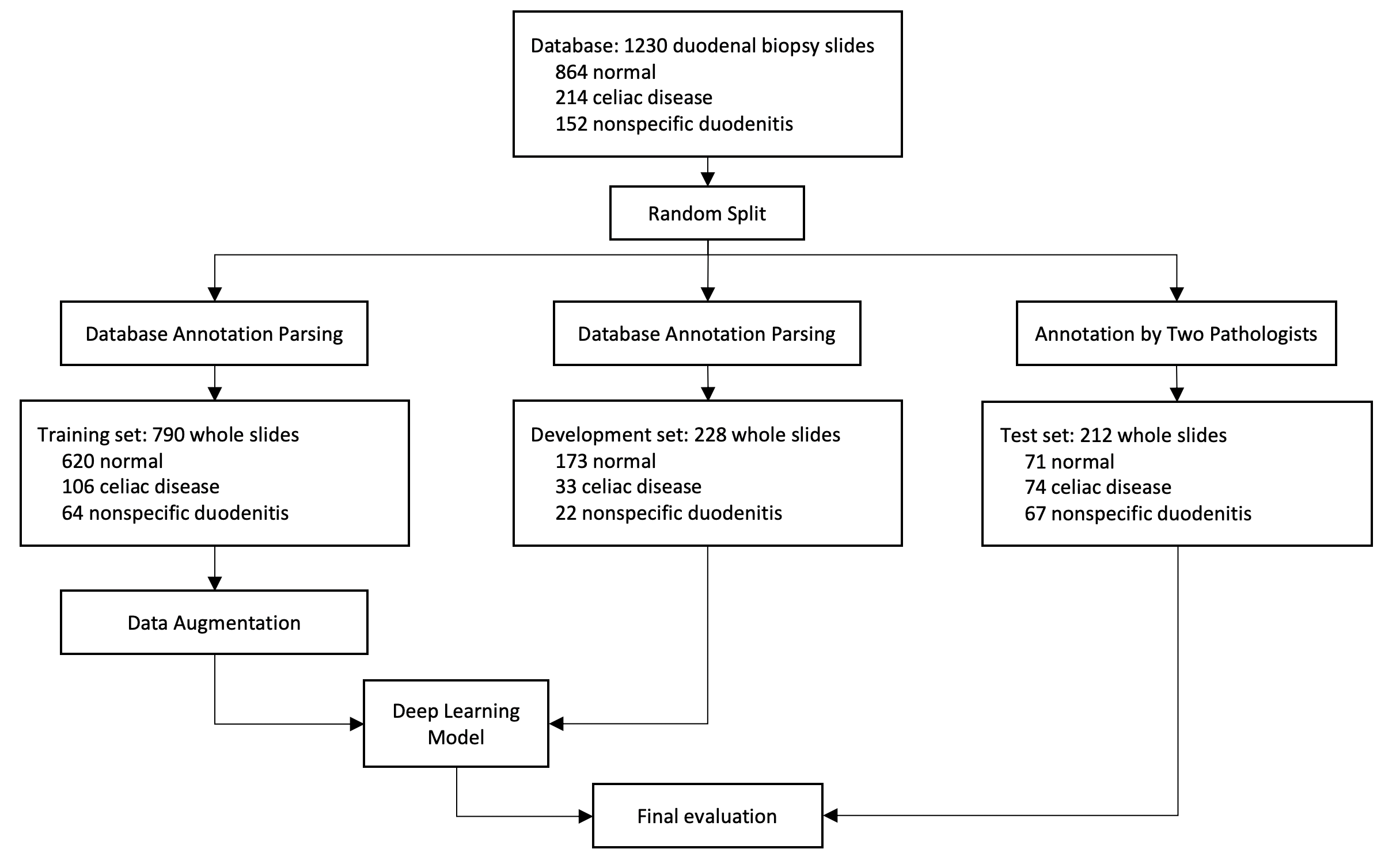}
\caption{Data flow diagram for allocating whole slides for training, development, and testing of our model. For training, patches were generated using the sliding window algorithm to train our ResNet patch classifier. The development set was used to fine-tune hyperparameters and thresholds of our neural network. Finally, we evaluated our model on the test set of 212 whole-slide images with reference labels.} 
\label{fig:pull}
\end{centering}
\end{figure*}

Recently, a subfield in artificial intelligence known as deep learning has produced a set of image analysis techniques that automatically extract relevant features, transforming the field of computer vision [23]. Deep neural networks use a data-driven approach to learn multi-level representations of data, allowing for comprehensive image analysis and classification [24]. These techniques are being increasingly applied to medical imaging to assist radiologists and pathologists [25]. In gastroenterology, previous studies have already used deep neural networks to classify colorectal polyps on biopsy and colonoscopy images [26-28], intraductal papillary mucinous neoplasms in MRI images [29], and diabetic retinopathy in retinal fundus photographs [30]. For CD in particular, large video datasets captured during endoscopies have facilitated quantitative analysis with deep learning [31, 32]. However, endoscopic classification is for the most part not used for confirming the diagnosis of CD. In this study, we developed a deep learning model that detects CD from duodenal biopsy images, the gold standard for diagnosis. We evaluated our model on an independent test set of 212 whole-slide images. 

\section{METHODS}

\subsection{Data Collection}
To train and evaluate our model for celiac disease detection, we collected whole-slide images from all patients who underwent duodenal biopsies from 2016 to 2018 at the Dartmouth-Hitchcock Medical Center (DHMC), a tertiary academic care center in Lebanon, NH. These slides contain hematoxylin-eosin stained formalin-fixed paraffin-embedded (FFPE) tissue specimens and were scanned by a Leica Aperio whole-slide scanner at 20x magnification by the Department of Pathology and Laboratory Medicine at DHMC. In total, we collected 1230 slides from 1048 patients. We randomly partitioned 1018 of these whole-slide images from 681 patients for model training and 212 whole-slide images from 163 patients as an independent test set for final evaluation of our model. There was no patient overlap for the slides in the training and test sets.

\subsection{Slide Annotation}
All whole-slide images used in our study were diagnosed by attending pathologists on gastrointestinal pathology service at the time as either normal, celiac disease, or nonspecific duodenitis. Normal duodenal biopsies show preserved villous architecture with no mucosal injury or acute or chronic inflammation. Celiac disease biopsies show a spectrum of histologic changes as described in Marsh classification [12], including partial to total villous atrophy with intraepithelial lymphocytosis, chronic inflammation and crypt regenerative hyperplasia. Nonspecific duodenitis includes histologic changes including peptic duodenitis, drug induced injury, and various other differential diagnoses of villous atrophy and acute and chronic inflammation. These labels were parsed from the medical record database and assigned as reference standard for slides used during model training. The training slides were then further split into a training set of 790 images and a development set of 228 images. Training set slides were used for training our neural network, while development set images were used for hyperparameter tuning. For the independent test set of 212 images, however, all labels were separately reviewed and confirmed by two gastrointestinal pathologists. Disagreements between original labels and new labels were reviewed by a senior gastrointestinal pathologist, who determined final classifications. The class distributions and roles in the data flow for our training, development, and test set are shown in Figure 1.

\subsection{Model Development}
In recent years, research in deep learning has demonstrated successful application of convolutional neural networks for image classification, including medical image analysis. In our study, we used the deep residual network (ResNet) [33], a neural network architecture built from residual blocks. ResNet significantly outperforms early deep learning models such as AlexNet [34] and VGG [35], and achieved state-of-the-art performance on the ImageNet and COCO image recognition benchmarks [36, 37]. We implemented ResNet to take in square patches as inputs and output a prediction probability for each of the three classes: normal, celiac disease, and nonspecific duodenitis.

For model training, we used a sliding window method on each high-resolution whole-slide image to generate small patches of size 224 by 224 pixels. Since some classes had more whole-slide images than others, we generated patches with different overlapping areas for each class. When inputting a patch into the model for training, we normalized the RGB color channels to the mean and standard deviation of the entire training set to neutralize differences in color among slides. Then, we performed color jittering on the brightness, contrast, saturation, and hue of each patch. Finally, we randomly rotated and flipped the images across the horizontal and vertical axes. In total, we generated 80,000 patches for each of our three classes, which were then uniquely augmented during each epoch in training.

In terms of model parameters, we initialized ResNet-50, the fastest ResNet with three-layer residual blocks, with weights from the He initialization [38]. We trained our ResNet model by optimizing on a multi-class cross-entropy loss function for forty epochs on the augmented training set, starting with an initial learning rate of 0.0001 and decaying by a factor of 0.85 every epoch. We used the Adam optimizer [39] and weight decay regularization (L2 penalty) [40] of 0.0001. Total training time was twelve hours on a Titan Xp graphics processing unit (GPU).

\subsection{Whole-Slide Inference}
In whole-slide inference, we aimed to classify each whole-slide image as either normal, celiac disease, or nonspecific duodenitis. The model is trained to classify small patches rather than entire slides, so we again used the sliding window algorithm to break down each whole slide into a collection of patches, each overlapping by one-third area. Next, we applied our trained ResNet model to classify each patch, and we filtered out noise using thresholding to discard predictions of low confidence. Given the distribution of patch predictions, we used the following heuristic to determine the whole-slide class: if more than $\gamma$ patches were classified as nonspecific duodenitis, then the whole slide was classified as nonspecific duodenitis. Otherwise, the most commonly predicted class was chosen as the whole-slide prediction. Thresholds for filtering noise, as well as $\gamma$, were optimized by performing a grid search over the development set. This allowed for accurate classification of slides with a significant amount of nonspecific duodenitis that was not covering the majority of the specimen area. Figure 2 depicts the whole-slide inference process. Inference time for a single whole-slide image was about fifteen seconds on a single Titan Xp GPU.

\begin{figure*}[ht]
\begin{centering}
\includegraphics[width=\linewidth]{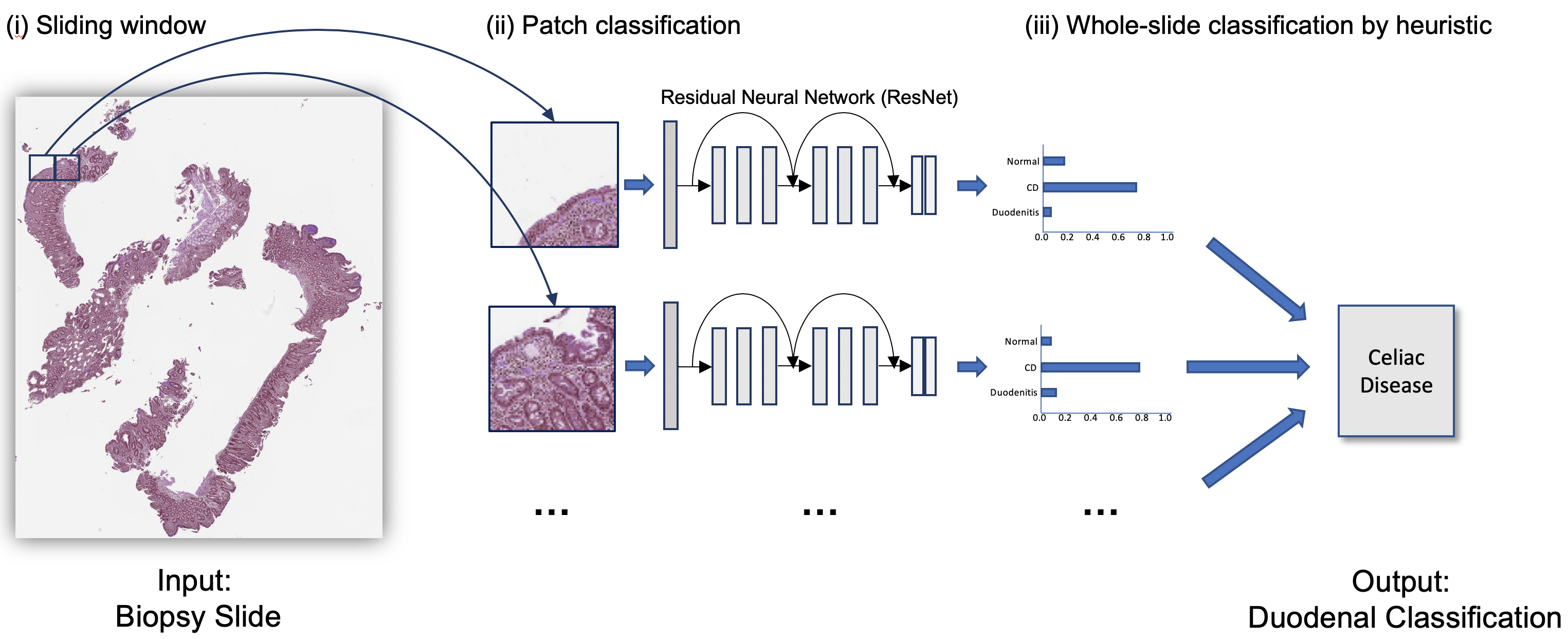}
\caption{Overview of detection of celiac disease on whole-slide biopsy images. We used a sliding window approach on a whole-slide image to generate patches, classified each patch with a ResNet model, and used a heuristic on the aggregated patch predictions to classify the whole slide.} 
\label{fig:pull}
\end{centering}
\end{figure*}

\subsection{Evaluation and Visualization}
For final evaluation, we applied our model to the independent test set of 212 whole-slide images. We compared the predictions of our model with reference standards established by pathologists, and measured accuracy, precision, recall, and F1 score for each class. We calculated confidence intervals for all performance metrics using the Clopper and Pearson method [41]. In addition, we plotted Receiver Operating Characteristics (ROC) curves and calculated area under the curve (AUC) for each class.

Furthermore, we visualized our model’s predictions at both the whole-slide and patch level. At the whole-slide level, we overlaid color-coded dots on patches for which the model predicted a particular pattern. This helps pathologists quickly identify regions of the slide containing abnormal tissue. At the patch level, we used the class activation mapping (CAM) method [42] to generate a pixel-level heat map that highlights the most informative regions of the image relevant to the predicted class. This demystified our classification method for each patch by revealing the most significant histologic features on the patch for each class for our model.

\section{RESULTS}
For model selection, we validated our neural network model on the development set of 228 images. We found the optimal thresholds for filtering out noise at the patch-level to be 0.7 for the normal class, 0.8 for the celiac disease class, and 0.85 for nonspecific duodenitis. For selection of the $\gamma$ threshold for percent area needed to classify a whole-slide as nonspecific duodenitis, we considered our gastrointestinal pathologist’s subjective examination of our model’s predictions in addition to a grid search to arrive at $\gamma$ = 0.25. After threshold optimization, our best model applied to the development set achieved an accuracy of 95.6\% for normal, 98.7\% for celiac disease, and 94.3\% for nonspecific duodenitis after threshold optimization. 

Performance of our model on the independent test is shown in Table 1, which includes accuracy, precision, recall, and F1 score with 95\% confidence intervals. Notably, our model detects the presence of CD with an accuracy of 95.3\% and an F1 score of 93.5\%. Table 2 shows the confusion matrix for predicted labels versus reference labels. ROC curves and AUC for each class are shown in Figure 3. AUC was greater than 0.95 for all classes. Figure 4 depicts whole-slide visualizations of twelve biopsy samples using dots to indicate predicted patch labels. Finally, CAM visualizations of individual patches are shown in Figure 5 to highlight relevant features used in our model’s classification process. A subjective qualitative investigation of these visualizations by a gastrointestinal pathologist confirmed that the predictions of our model are generally on target.

\begin{table*}[t]
\centering
\begin{tabular}{lcccc}
\hline
& Accuracy (\%) &	Precision (\%)	& Recall (\%) &	F1 Score (\%) \\ \hline
Normal (n = 71) &	91.0 (87.2-94.9)&	83.3 (75.1-91.6)&	91.5 (85.1-98.0) &	87.2 (79.9-95.2)\\
Celiac Disease (n = 74)	&95.3 (92.4-98.1)&	90.0 (83.4-96.6)&	97.3 (93.6-99.9)&	93.5 (87.8-99.3) \\
Nonspecific Duodenitis (n = 67)	&89.2 (85.0-93.3)&	90.7 (83.0-98.5)&	73.1 (62.5-83.7)&	81.0 (71.5-90.5) \\ \hline
Average&	87.7 (83.3-92.2)&	88.0 (80.5-95.4)&	87.3 (79.5-95.2)&	87.2 (79.4-95.1)\\ \hline
\end{tabular}
\caption{Performance of our final model for celiac disease detection on 212 duodenal biopsy whole-slide images in our test set. 95\% confidence intervals are shown in parentheses.}
\end{table*}

\begin{table}[ht]
\begin{tabular}{l c c c}
\hline
Prediction & \multicolumn{3}{c}{Reference}\\ \hline \hline
 & Normal& 	Celiac Disease& 	Duodenitis \\ \hline
 Normal& 	65& 	2& 	11 \\
Celiac Disease& 	1& 	72& 	7 \\
Duodenitis& 	5& 	0& 	49 \\ \hline
\end{tabular}
\caption{Confusion matrix of our final model for celiac disease detection on 212 duodenal biopsy whole-slide images in our test set.}
\label{tab:data}
\end{table}

\begin{figure}[ht]
\begin{centering}
\includegraphics[width=\linewidth]{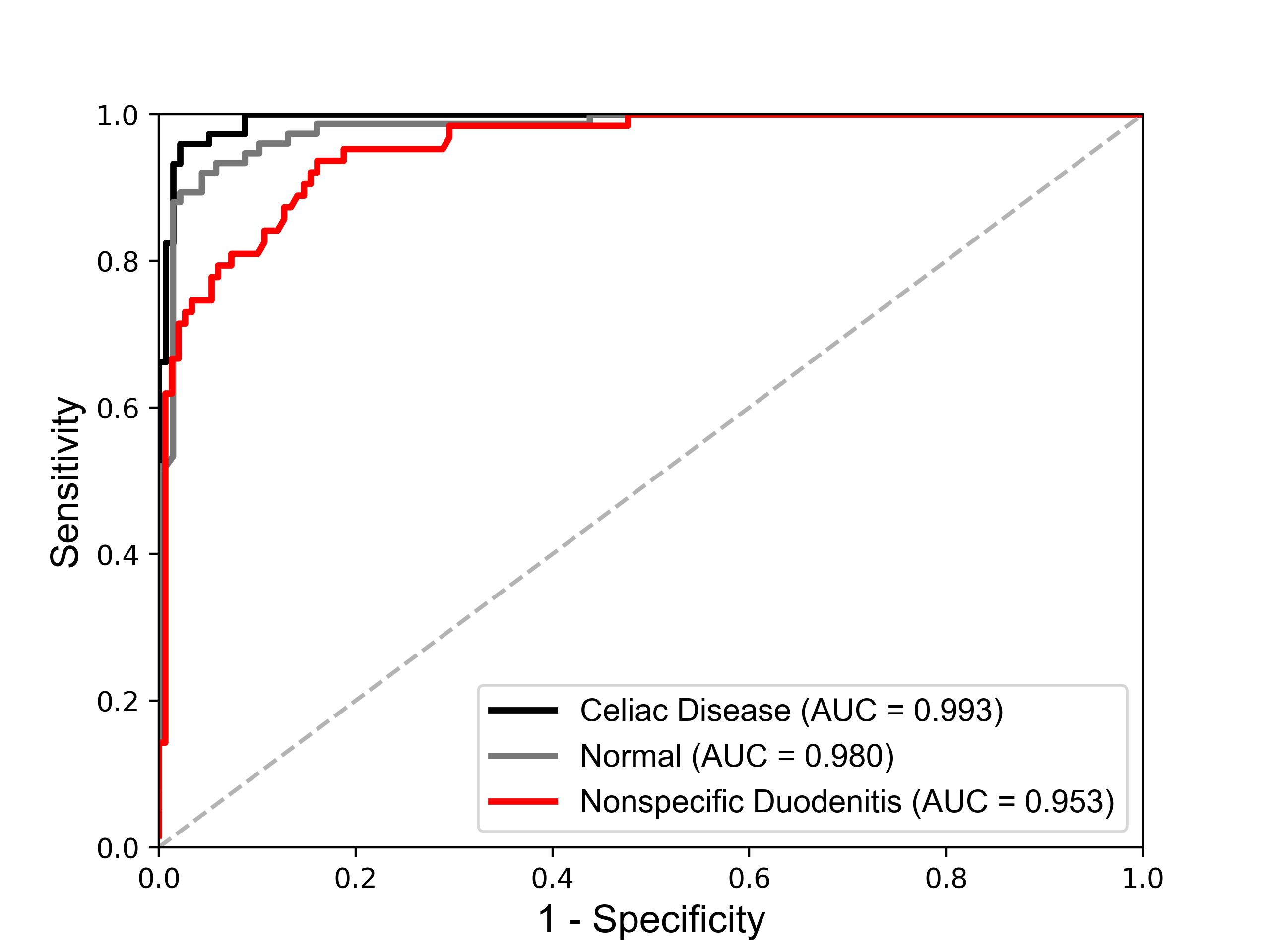}
\caption{ROC curves and their area under the curves (AUC’s) for our model’s classifications on the independent test set of 212 whole-slide biopsy images.} 
\label{fig:pull}
\end{centering}
\end{figure}

\begin{figure*}[tbp]
\begin{centering}
\includegraphics[width=\linewidth]{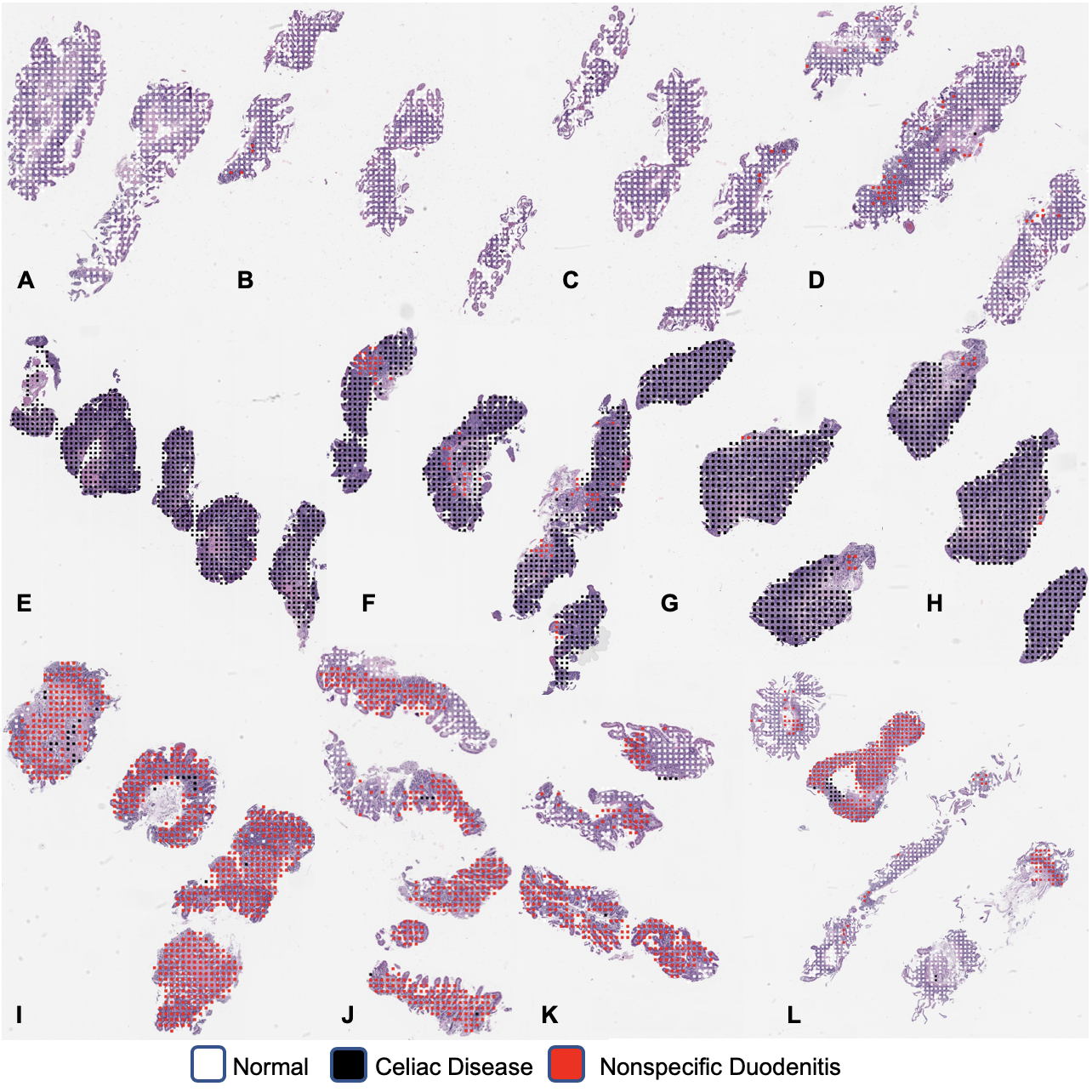}
\caption{Visualization of patch predictions of our model at the whole-slide level. A-D was correctly classified as normal, E-H was correctly classified as celiac disease, and I-L was correctly classified as nonspecific duodenitis.} 
\label{fig:pull}
\end{centering}
\end{figure*}

\section{DISCUSSION}
Duodenal biopsies are the gold standard for confirming the diagnosis of CD. The prevalence of CD has increased dramatically in recent years, and active case-finding calls for more serological tests and duodenal biopsies. Detection of CD on these biopsies could potentially be enhanced and facilitated by automated image processing. In this study, we presented a deep learning model that classifies duodenal tissue and highlights the associated features and regions of interest. Previous work has used deep learning to detect CD from endoscopic images [31, 32]. While acknowledging the substantive work of these investigators, endoscopies are for the most part not used for confirming CD diagnoses. Our model not only detects CD on duodenal biopsies, but it also visualizes regions of normal tissue, celiac disease, and nonspecific duodenitis to aid review by pathologists. Of note, we are not aware of any other existing system for CD detection on biopsy images.

Our model achieved high performance for detection of celiac disease. On the independent test set of 212 images, our model detected CD with a considerable F1 score of 93.5\% and AUC of 0.993. Since our model made predictions at the patch level and then aggregated them for whole-slide inference, it was relatively unaffected by noise and achieved high accuracy. For normal and nonspecific duodenitis, F1 scores were 87.2\% and 81.0\% respectively. Identification of nonspecific duodenitis was more challenging because this class’s tissue often also contains some normal tissue fragments, which complicate the analysis. Sixteen slides were misclassified between normal and nonspecific duodenitis, and pathologist evaluation of these errors revealed errors related to tissue orientation, fixation artifact, and patchy histologic changes. In addition, seven slides of nonspecific duodenitis were identified as celiac disease due to focal increase of intraepithelial lymphocytes and partial villous atrophy. Performance measures for the nonspecific duodenitis class were the lowest across the board. There are several reasons for this. One could be that nonspecific duodenitis had the lowest number of training samples, comprising only 64 images in the training set compared to 620 and 106 for normal tissue and celiac disease respectively. Another could be that this category comprises several disease entities including peptic chronic duodenitis, active duodenitis, and other nonspecific reactive changes, making it more challenging to detect since there was a wider range of histologic attributes to learn. Finally, slides labeled as nonspecific duodenitis often contained some portions of normal tissue, and since we extracted patch labels based on whole-slide labels during the training process, it is likely that some mislabeled data was used in model training and made it harder to detect this class in our approach.

In terms of visualization, the class activation mapping results of our model’s selected areas of attention indicate that our model has learned the correct histologic features for each class. Classification of normal tissue tended to be holistic, with attention to almost all tissue area including normal villous architecture. For celiac disease, the model correctly identified villous atrophy, intraepithelial lymphocytosis, and chronic inflammation in the lamina propria. In the case of nonspecific duodenitis, the model identified villous thickening, Brunner’s gland hyperplasia, foveolar metaplasia, and chronic inflammation.

\begin{figure*}[tbp]
\begin{centering}
\includegraphics[width=\linewidth]{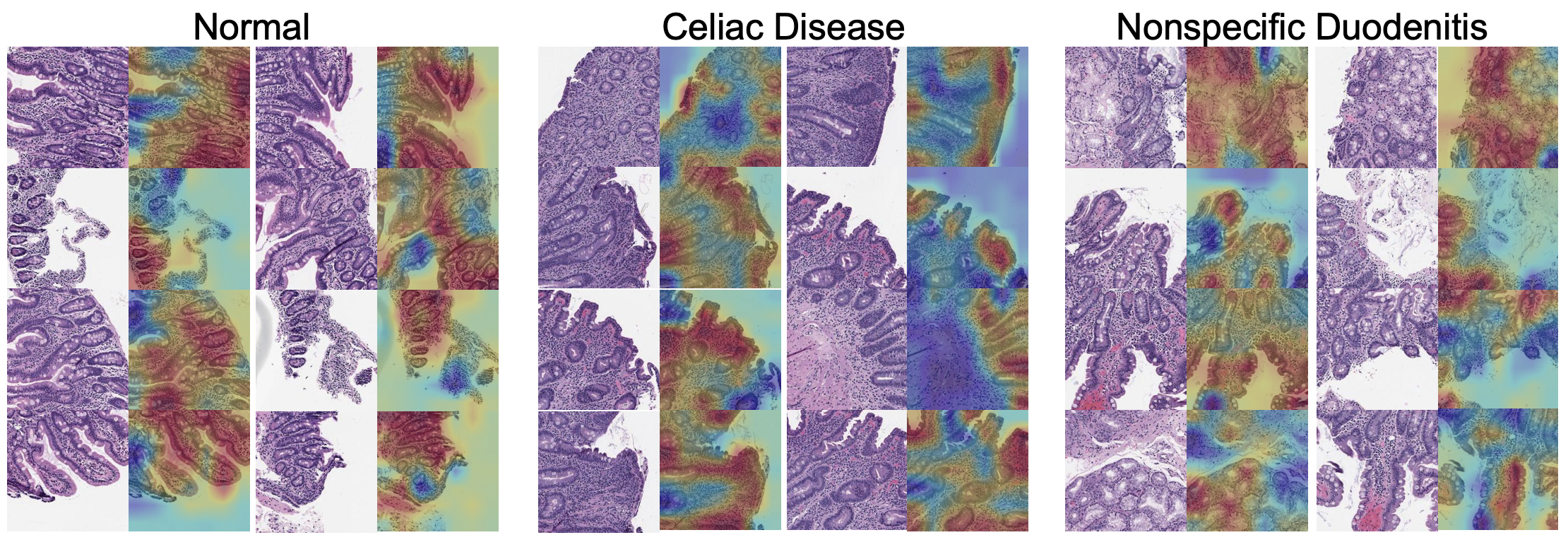}
\caption{Class activation mapping (CAM) heat maps highlighting most informative regions of patches relevant to normal, celiac disease, and nonspecific duodenitis classes. Red regions indicate areas of attention for our residual neural network. Classification of normal tissue tends to be holistic, with attention to almost all tissue area including normal villous architecture. For celiac disease, the model correctly identified villous atrophy, intraepithelial lymphocytosis, and chronic inflammation in the lamina propria. In the case of nonspecific duodenitis, the model identified villous thickening, Brunner’s gland hyperplasia, foveolar metaplasia, and chronic inflammation.} 
\label{fig:pull}
\end{centering}
\end{figure*}

Our results indicate that deep neural networks have substantial potential to aid gastrointestinal pathologists in diagnosing CD. For application in a clinical setting, our model could be integrated into existing laboratory information management systems to pre-populate patch predictions on slides and provide preliminary diagnoses prior to review by pathologists. In addition, a visualization of the slide evaluated by our model at the piecewise level could highlight precise tissue area containing abnormal or sprue patterns, allowing pathologists to quickly examine regions of interest. As CD prevalence has increased dramatically in recent years, more serological screenings and duodenal biopsies are being done for patients at risk [21, 22]. With biopsies as the gold standard for diagnosis, our work aims to provide pathologists with a tool for more accurate and efficient detection of CD. 

The model presented in this paper is rooted in solid deep learning methodology and achieves commendable performance, but there are several limitations of our study. One limitation is that all biopsy slides were collected from a single medical center and scanned with the same equipment, so our data may not be representative of the entire range of histologic patterns in patients worldwide. Although our whole-slide scans are high resolution and we were able to extract a large number of patches for training with the sliding window method, our dataset is still small in comparison to conventional datasets in deep learning, which contain more than ten thousand unique samples per class [43, 44] and more than a million unique images in total [45, 46]. Overfitting is unlikely because we generated a large number of small patches for training and conducted final evaluation on an independent test set, but it is still a possibility. Collecting more data in collaboration with another medical center in future work would allow us to train a more generalizable neural network and could also improve our model’s performance in classifying nonspecific duodenitis.

Moving forward, more work will be done to further the capabilities of our model and evaluate its use a clinical setting. Collecting an annotated dataset with specific histopathological classifications of celiac disease and labeled bounding boxes around lesions would allow our model to classify and locate specific lesion types, providing pathologists with more comprehensive slide analysis, particularly for the nonspecific duodenitis class. Furthermore, once more data is collected, we can predict slide level results by using patch predictions to train a traditional machine learning classifier such as a support vector machine or random forest, which may yield better results than our current thresholding method. In terms of clinical application, we plan on validating our model on a larger test set from multiple institutions and deploying a trial implementation of our model into laboratory information management systems at the Dartmouth-Hitchcock Medical Center to measure its ability to improve CD detection accuracy and efficiency. However, widespread clinical implementation of such artificial intelligence tools will require major future steps, which our group will be undertaking. Any deep learning models for computer-aided pathology must be thoroughly validated through clinical trials and be proven to enhance outcomes. Such model must also not impact the established workflow of pathologists or slow down the speed of existing programs. Most importantly, deep learning models must be accurate and gain the confidence of physicians, patients, and the medical community. In its current state, artificial intelligence has the ability to analyze images and make preliminary classifications, but much more work must be done before patients and physicians will be able to trust computers to make medical decisions. We believe the work presented in this paper is a preliminary step in this direction.

In summary, we have demonstrated that deep learning can achieve high accuracy in detecting CD in duodenal biopsies. Our model uses a state-of-the-art residual neural network architecture for whole-slide classification and achieved exemplary results on an independent test set of 212 whole-slide images. As CD prevalence and screening increases, we expect our model could assist pathologists in more accurate and efficient evaluation of duodenal biopsy slides.

\section{ACKNOWLEDGEMENTS}
We thank Matthew Suriawinata for assistance with slide scanning and Sophie Montgomery and Lamar Moss for their feedback on the manuscript. This research was supported in part by a National Institutes of Health grant, P20GM104416.

\newpage
\section{REFERENCES}
\begin{enumerate}[label={[\arabic*]}]
\item  Green PH, Cellier C. Celiac disease. N Engl J Med. 2007;357:1731-1743.
\item  Mustalahti K, Catassi C, Reunanen A, et al. The prevalence of celiac disease in Europe: Results of a centralized, international mass screening project. Ann Med. 2010;42:587-595.
\item  Rubio-Tapia A, Hill ID, Kelly CP, et al. American College of Gastroenterology clinical guideline: diagnosis and management of celiac disease. Am J Gastroenterol. 2013;108:656-677.
\item  Rubio-Tapia A, Kyle RA, Kaplan EL, et al. Increased prevalence and mortality in undiagnosed celiac disease. Gastroenterology. 2009;137:88-93. 
\item  Godfrey JD, Brantner TL, Brinjikji W, et al. Morbidity and mortality among older individuals with undiagnosed celiac disease. Gastroenterology 2010;139:763-769.
\item  Dube C, Rostom A, Sy R, et al. The prevalence of celiac disease in average-risk and at-risk Western European populations: a systematic review. Gastroenterology 2005;128:57-67.
\item  West J, Logan RF, Hill PG, et al. Seroprevalence, correlates, and characteristics of undetected coeliac disease in England. Gut. 2003;52:960-965.
\item  Rostami K, Mulder CJJ, Werre JM, et al. High prevalence of celiac disease in apparently healthy blood donors suggests a high prevalence of undiagnosed celiac disease in the Dutch population. Scand J Gastroenterol. 1999;34:276-279.
\item  Rubio-Tapia A, Murray JA. Classification and management of refractory coeliac disease. Gut. 2010;59:547-557.
\item  Green PHR, Rostami K, Marsh M. Diagnosis of coeliac disease. Best Pract Res Clin Gastroenterol. 2005;19:389-400.
\item  Bryne G, Feighery CF. Celiac disease: diagnosis. Methods Molecular Biology, vol 1326. 2015.
\item  Marsh MN. Gluten, major histocompatibility complex, and the small intestine: a molecular and immunobiotic approach to the spectrum of gluten sensitivity (‘celiac sprue’). Gastroenterology. 1992;102:330-354.
\item  Oberhuber G. Histopathology of celiac disease. Biomed Pharmacother. 2000; 54:368-372.
\item  Corazza GR, Villanacci V, Zambelli C, et al. Comparison of the interobserver reproducibility with different histologic criteria used in celiac disease. Clin Gastroenterol Hepatol. 2007;5:838-843.
\item  Montgomery EA, Voltaggio L. Biopsy Interpretation of the Gastrointestinal Tract Mucosa: Volume 1. 2011. 
\item  Fasano A, Catassi C. Current approaches to diagnosis and treatment of celiac disease: an evolving spectrum. Gastroenterology. 2001;120:636-651.
\item  Corazza G, Villanacci V, Zambelli C, et al. Comparison of the interobserver reproducibility with different histologic criteria used in celiac disease. CGH. 2007;5:838-843.
\item  Mubarak A, Nikkels P, Houwen R, et al. Reproducibility of the histological diagnosis of celiac disease. Scand J Gastroenterol. 2011;46:1065-1073.
\item  Arguelles-Grande C, Tennyson C, Lewis S, et al. Variability in small bowel histopathology reporting between different pathology practice settings: impact on the diagnosis of coeliac disease. J Clin Pathol. 2012;65:242-247.
\item  Taavela J, Koskinen O, Huhtala M, et al. Validation of morphometric analyses of small-intestinal biopsy readouts in celiac disease. PLoS One. 2013:8(10).
\item  Lacucci M, Ghosh S. Routine duodenal biopsies to diagnose celiac disease. Ca J Gastroenterol. 2013;27:385.
\item  Serra S, Jani PA. An approach to duodenal biopsies. J Clin Pathol. 2006;59:1133-1150.
\item  Lecun Y, Bengio Y, Hinton G. Deep learning. Nature. 2015;521:436.
\item	Goodfellow I, Bengio Y, Courville A. Deep learning. The MIT Press. 2016.
\item	Tomita N, Cheung YY, Hassanpour S. Deep neural networks for automatic detection of osteoporotic vertebral fractures on CT scans. Comp Bio Med. 2018;98:8-15.
\item	Korbar B, Olofson AM, Miraflor AP, et al. Looking Under the Hood: Deep Neural Network Visualization to Interpret Whole-Slide Image Analysis Outcomes for Colorectal Polyps. CVPR Workshops. 2017;69-75. 
\item  Byrne MF, Chapados N, Soudan F, et al. Real-time differentiation of adenomatous and hyperplastic diminutive colorectal polyps during analysis of unaltered videos of standard colonoscopy using a deep learning model. Gut. 2017;0:1-7.
\item  Korbar B, Olofson AM, Miraflor AP, et al. Deep learning for classification of colorectal polyps on whole-slide images. J Pathol Inform. 2017;8:30.
\item  Corral J, Hussein S, Kandel P, et al. Deep learning to diagnose intraductal papillary mucinous neoplasms (IPMN) with MRI. Gastroenterology. 2018;154:S524.
\item  Gulshan V, Peng L, Coram M, et al. Development and validation of a deep learning algorithm for detection of diabetic retinopathy in retinal fundus photographs. JAMA. 2016;314:2402-2410.
\item  Zhou T, Han G, Li B, et al. Quantitative analysis of patients with celiac disease by video capsule endoscopy: a deep learning method. Comp Bio Med. 2017;85:1-6.
\item  Gadermayr M, Wimmer G, Kogler H, et al. Automated classification of celiac disease during upper endoscopy: Status quo and quo vadis. Comp Bio Med. 2018;102:221-226.
\item  He K, Zhang X, Ren S,  et al. Deep residual learning for image recognition. CVPR. 2016;770-778.
\item  Krizhevsky A, Sutskever I, Hinton G. ImageNet Classification with deep convolutional neural networks. NIPS. 2012;1097-1105.
\item  Simonyan K, Zisserman A. Very deep convolutional networks for large-scale image recognition. ICLR. 2015.
\item 	Russakovsky O, Deng J, Hao S, et al. ImageNet large scale visual recognition challenge. IJCV. 2015;115:211-252.
\item  Lin TY, Maira M, Belongie S, et al. Microsoft COCO: common objects in context. ECCV. 2014;8693:740-755.
\item  He K, Zhang X, Ren S, et al. Delving deep into rectifiers: surpassing human-level performance on ImageNet classification. ICCV. 2015;1026-1034.
\item  Kingma DP, Ba J. Adam: a method for stochastic optimization. ICLR. 2015.
\item  Krogh A, Hertz JA. A simple weight decay can improve generalization. NIPS. 1991;950-957.
\item  Clopper CJ, Pearson ES. The use of confidence or fiducial limits illustrated in the case of the binomial. Biometrika 1934;26:404-413.
\item  Zhou B, Khosla A, Lapedriza A, et al. Learning deep features for discriminative localization. CVPR. 2016:2921-2929.
\item  LeCun Y, Cortes C. MNIST handwritten digit database. 2010. Available from https://yann.lecun.com/exdb/mnist.
\item  Netzer Y, Wang T, Coates A, et al. Reading Digits in Natural Images with Unsupervised Feature Learning. NIPS. 2011.
\item  Russakovsky O, Deng J, Su H, et al. ImageNet Large Scale Visual Recognition Challenge. IJCV. 2015;115:211-252.
\item  Krasin I, Duerig T, Alldrin N, et al. OpenImages: a public dataset for large-scale multi-label and multi-class image classification. 2017. Available from https://storage.googleapis.com/openimages/web/index.html.

\end{enumerate}



\end{document}